\documentclass[times,twocolumn,final,authoryear]{elsarticle}

\usepackage{ycviu}
\usepackage{framed,multirow}

\usepackage{latexsym}

\usepackage{url}
\usepackage{xcolor}
\definecolor{newcolor}{rgb}{.8,.349,.1}

\usepackage{graphicx}
\usepackage{comment}
\usepackage{textcomp}
\usepackage{amsmath,amssymb} 
\usepackage{color}
\usepackage{gensymb}
\usepackage{graphicx}
\usepackage{rotating}
\usepackage{color}
\usepackage[caption=false]{subfig}
\usepackage{framed}
\usepackage{balance}
\usepackage{booktabs}
\usepackage{bm}
\usepackage{bbm}
\usepackage{enumitem}
\usepackage[export]{adjustbox}
\usepackage{epigraph}
\usepackage{hyperref}

\newcommand{\ie}{\textit{i}.\textit{e}., }
\newcommand{\eg}{\textit{e}.\textit{g}. }
\newcommand{\tit}[1]{\medbreak\noindent\textbf{#1.}}

\newcommand{\tinytit}[1]{\noindent\textbf{#1.}}

\newcommand{\rev}[1]{\textcolor{black}{#1}}
\newcommand{\revmin}[1]{\textcolor{black}{#1}}

\journal{Computer Vision and Image Understanding}

\begin{document}

\thispagestyle{empty}

\begin{frontmatter}

\title{Multimodal Attention Networks for Low-Level Vision-and-Language Navigation}

\author[1]{Federico \snm{Landi}\corref{cor1}} 
\cortext[cor1]{Corresponding author:}
\ead{federico.landi@unimore.it}
\author[1]{Lorenzo \snm{Baraldi}}
\author[1]{Marcella \snm{Cornia}}
\author[2]{Massimiliano \snm{Corsini}}
\author[1]{Rita \snm{Cucchiara}}

\address[1]{University of Modena and Reggio Emilia, Italy}
\address[2]{ISTI - CNR, Italy}

\received{1 May 2013}
\finalform{10 May 2013}
\accepted{13 May 2013}
\availableonline{15 May 2013}
\communicated{S. Sarkar}

\begin{abstract}
Vision-and-Language Navigation (VLN) is a challenging task in which an agent needs to follow a language-specified path to reach a target destination. 
The goal gets even harder as the actions available to the agent get simpler and move towards low-level, atomic interactions with the environment. This setting takes the name of low-level VLN.
In this paper, we strive for the creation of an agent able to tackle three key issues: multi-modality, long-term dependencies, and adaptability towards different locomotive settings. To that end, we devise ``Perceive, Transform, and Act'' (PTA): a fully-attentive VLN architecture that leaves the recurrent approach behind and the first Transformer-like architecture incorporating three different modalities -- natural language, images, and low-level actions for the agent control.
In particular, we adopt an early fusion strategy to merge lingual and visual information efficiently in our encoder. We then propose to refine the decoding phase with a late fusion extension between the agent's history of actions and the perceptual modalities.
\revmin{We experimentally validate our model on two datasets: PTA achieves promising results in low-level VLN on R2R and achieves good performance in the recently proposed R4R benchmark.}
Our code is publicly available at \url{https://github.com/aimagelab/perceive-transform-and-act}.
\end{abstract}



\end{frontmatter}


\section{Introduction}
\label{sec:intro}
%
Effective instruction-following and contextual decision-making can open the door to a new world for researchers in embodied AI.  Deep neural networks have the potential to build complex reasoning rules that enable the creation of intelligent agents, and research on this subject could also help to empower the next generation of collaborative robots~\citep{Savva_2019_ICCV,xia2018gibson}.
In this scenario, Vision-and-Language Navigation (VLN)~\citep{anderson2018vision} plays a significant part in current research. This task requires to follow natural language instructions through unknown environments, discovering the correspondences between lingual and visual perception step by step.
Additionally, the agent needs to progressively adjust navigation in light of the history of past actions and explored areas. Even a small error while planning the next move can lead to failure because perception and actions are unavoidably entangled; indeed, \textit{we must perceive in order to move, but we must also move in order to perceive}~\citep{gibson2014ecological}. For this reason, the agent can succeed in this task only by efficiently combining the three modalities -- language, vision, and actions.

\begin{figure*}[t!]
\centering
\subfloat[Low-level and High-level action spaces]%
{\includegraphics[height=0.3\linewidth,valign=b]{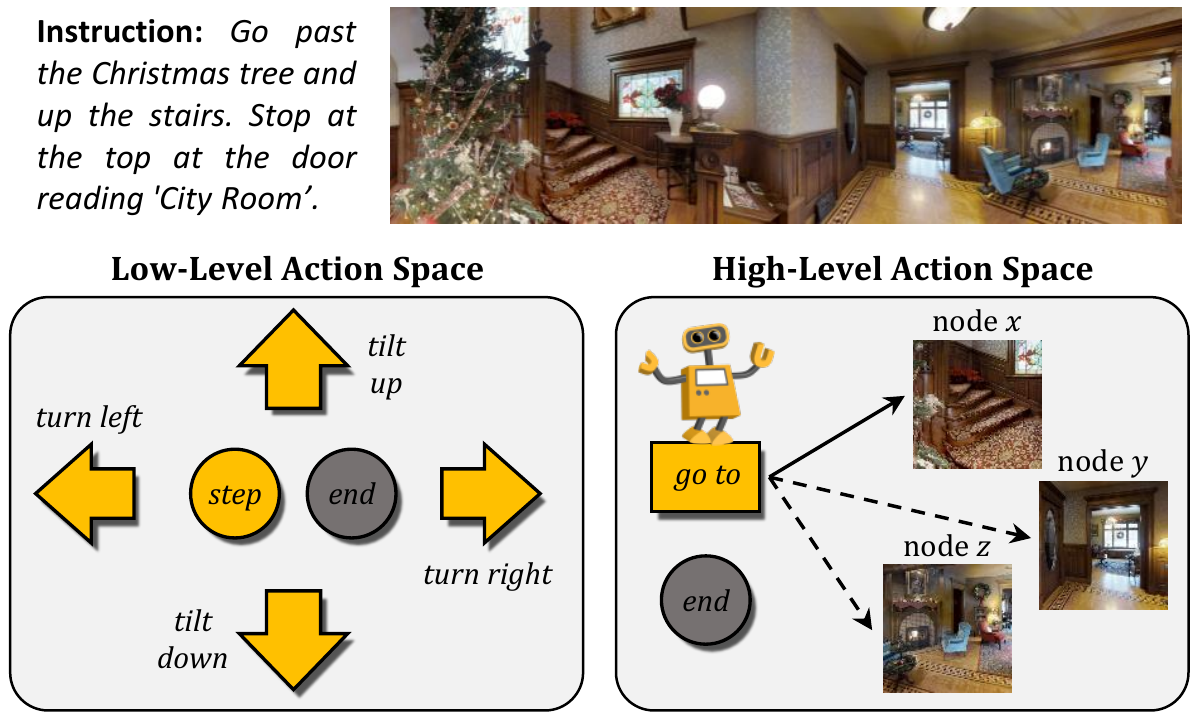}%
\label{fig:Figure01_a}}
\hspace{0.4cm}
\subfloat[PTA is entirely based on attention]%
{\includegraphics[height=0.3\linewidth,valign=b]{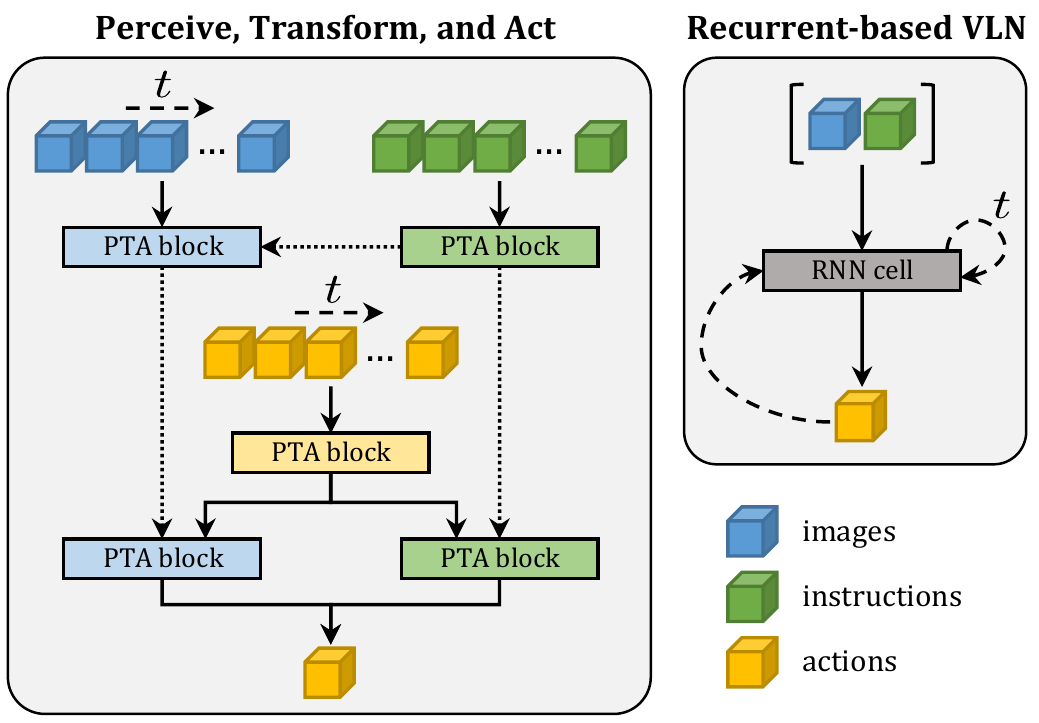}%
\label{fig:Figure01_b}}
\caption{\revmin{Previous approaches to VLN perform high-level navigation, relaxing the assumptions on the agent action space, and build upon recurrent neural networks to model long-term dependencies among the three modalities involved -- text, images, and actions.} Instead, PTA implements low-level interactions with the environment and exploit only attention mechanisms}
    \label{fig:Figure01}
    \vspace{-0.4cm}
\end{figure*}

Recent literature identifies two main operating settings for VLN~\citep{landi2019embodied}, called \textit{high-level action space} and \textit{low-level action space}~(Fig.~\ref{fig:Figure01_a}).
The concept of a high-level, \textit{panoramic} action space was first proposed by \cite{fried2018speaker}. In this setting, navigation takes place on a graph whose connectivity is known a priori and the nodes are represented by different viewpoints (\ie the locations where the agent can step and look at the surroundings). High-level agents predict the path to the goal as a sequence of connected viewpoints, and move through the environment using a teleporting system. This aspect limits adaptability to real-world applications and prevents current research on high-level VLN from having a practical impact on embodied navigation robots.
Instead, low-level methods make predictions over the agent locomotor system, hence performing actions with a one-to-one correspondence with the robot control system -- \textit{rotate $X\degree$}, \textit{tilt up/down}, and \textit{step forward} are examples of low-level actions. Even though low-level navigation can still be performed on a graph-like environment (with viewpoints as nodes), the agent is not aware of it and does not exploit any knowledge related to the structure of the underlying simulating platform. This setting is more in line with recent research on embodied AI platforms~\citep{Savva_2019_ICCV,xia2018gibson}, which is moving towards realistic and low-level interactions with the environment and continuous control of the agent.
Since the adaptability to real-world applications represents an important challenge in this scenario, we tackle the task of low-level VLN, in which abstract reasoning (\ie teleporting from a viewpoint to the next and knowledge of the connectivity graph) is no longer available to the agent.

Encouraged by the success of attention in many vision-and-language tasks~\citep{devlin2018bert,lu2019vilbert,vaswani2017attention}, we propose a new model for low-level VLN that exploits fully-attentive networks to merge the knowledge coming from different domains. In this work, we devise \textit{Perceive, Transform, and Act} (PTA), in which the different modalities (text, vision, and actions) can be conditioned on the full history of previous observations. 
While all the previous approaches to VLN rely on a recurrent policy to track the agent's internal status through time, we directly infer the state from the observations via attention and avoid any form of recurrence~(Fig.~\ref{fig:Figure01_b}). For this reason, our agent can model the dependencies tied to navigation more efficiently and generalize to longer episodes better than other models.

At the present time, there is no study exploring the possibility for a given architecture to switch between the high-level and the low-level action spaces. In this work, we experimentally show that methods born and designed for high-level navigation experience a drop in performance when adapted for low-level VLN. Indeed, high-level reasoning and abstraction from the physical environment is too heavily exploited to let the agent walk on its own. This is not true for PTA, which is designed for low-level use but can easily adapt to high-level scenarios.
%
We summarize our main contributions as follows:
\begin{itemize}
\setlength\itemsep{0.05em}
    \item We propose a novel multimodal framework for low-level VLN that replaces any form of recurrence with attention mechanisms, using them to tackle both long-term dependencies and multi-modality. To the best of our knowledge, our model is the first Transformer-like architecture to merge visuo-linguistic perception with information coming from the agent action system;
    \item We technically describe how it is possible to switch from a high-level output space to a low-level locomotor system and vice versa. \revmin{Experimental results on this subject are the first to analyze the mutual relationships between low-level and high-level VLN, and validate the hypothesis that high-level architectures are not easily adaptable to the low-level counterpart. Such results highlight the need for more experiments in this direction for future works;}
    \item \revmin{Experimental results show that PTA achieves good performance on low-level VLN.} We validate this claim on two different benchmarks of increasing instruction length and complexity. Since our setting is closer to real-world applications and requires to decode fine-grained atomic actions, we believe that low-level VLN represents the next testbed for embodied agents aiming to perform Vision-and-Language Navigation.
\end{itemize}

\section{Related Work}
\label{sec:related}
There is a wide area of research devoted to bridge natural language processing and image understanding. Image captioning~\citep{Anderson_2018,vinyals2015show,xu2015show}, visual question answering~\citep{VQA,balanced_vqa_v2}, and visual dialog~\citep{visdial,visdial_rl} are examples of active research areas in this field. At the same time, visual navigation~\citep{Gupta2019cognitivemapper,Shen2019fusion,xia2018gibson} and goal-oriented instruction following~\citep{chen2019touchdown,fu2018from,qi2019rerere} represent an important part of current work on embodied AI~\citep{das2018embodied,das2018neural,Savva_2019_ICCV,yang2019embodied}.
In this context, Vision-and-Language Navigation (VLN)~\citep{anderson2018vision} constitutes a peculiar challenge, as it enriches traditional navigation with a set of visually rich environments and detailed instructions. Additionally, all the scenes are photo-realistic and unknown to the agent beforehand.

\tit{High-level Vision-and-Language Navigation}
The idea of a high-level action space was first proposed by \cite{fried2018speaker}, and immediately allowed for an important boost in terms of performance. Following work includes visual and textual co-grounding with progress inference~\citep{ma2019self} and backtracking with learned heuristics~\citep{ma2019regretful}. Other methods implement a speaker module which strengthens consistency between the chosen path and the instruction~\citep{fried2018speaker,wang2018reinforced}. \cite{wang2018reinforced} propose a reinforced cross-modal matching critic, together with a new self-supervised imitation learning setting. \cite{tan2019learning} devise a novel environmental dropout method to improve traditional features dropout for VLN. \cite{ke2017tactile} propose a FAST navigation agent which improves the performance both over greedy decoding of the next action and over beam search.
Very recently, \cite{zhu2019vision} exploit auxiliary reasoning tasks and the rich semantic given by the navigation in their model, while \cite{hao2020towards} investigate an efficient pre-training for generic VLN agents.
While pragmatic approaches with high-level reasoning allow for a boost in performance, architectures built for high-level VLN rely heavily on the information coming from the underlying simulating platform. Even when the environment is supposed to be unknown (\eg during test) the agent can get a priori knowledge from the connectivity graph and exploit this information for a more efficient navigation.  
Recently proposed benchmarks and new evaluation metrics~\citep{sotp2019acl} show that traditional approaches hardly adapt to longer trajectories. Indeed, the recurrent nature of previous methods exacerbates the difficulty of learning long-term dependencies~\citep{bengio1994learning} both in the instruction and in the navigation.

\rev{After the initial submission of this paper, new methods have been proposed to deal with VLN on a high-level perspective: a recent line of work designs graph operations to boost planning capabilities~\citep{deng2020evolving} or to model visuo-linguistic relationships in the graph nodes~\citep{hong2020language}. \cite{zhang2020language} propose to employ two levels of attention-guided co-grounding, together with a new learning scheme alternating teacher-forcing and student-forcing. \cite{qi2020object} design an architecture taking advantage from both visual tokens and action tokens in the instructions. Visual tokens are employed to identify meaningful visual features in the environment, while action tokens consider only the agent state (represented by coordinates features). In this work, we leverage the same intuition in our multi-modal decoder. In fact, we propose an additional decoding branch that does not employ visual features, but focuses on action clues provided in the sole instruction.}

\tit{Low-level Vision-and-Language Navigation} 
In low-level VLN, the agent takes move in the environment by using actions such as \textit{rotate}, \textit{tilt up}, and \textit{step ahead}. So far, only a small portion of literature has taken this direction. \cite{anderson2018vision} build on a traditional sequence-to-sequence architecture, while \cite{wang2018look} employ a mixture of model-free and model-based reinforcement learning. In these works the agent perceives only the first person view of the surrounding environment. More recently, \cite{landi2019embodied} propose a sequence-to-sequence model which exploits dynamic convolution to make the visual representation more compact and informative for the agent. In this last work, the agent perceives the 360\degree~image of the surroundings. This generalization does not hurt adaptability to real-world scenarios, since it is relatively easy to enrich the agent with additional RGB cameras. 

\section{Perceive, Transform, and Act}
\label{sec:method}
\begin{figure*}[t!]
\centering
    \includegraphics[width=.98\linewidth]{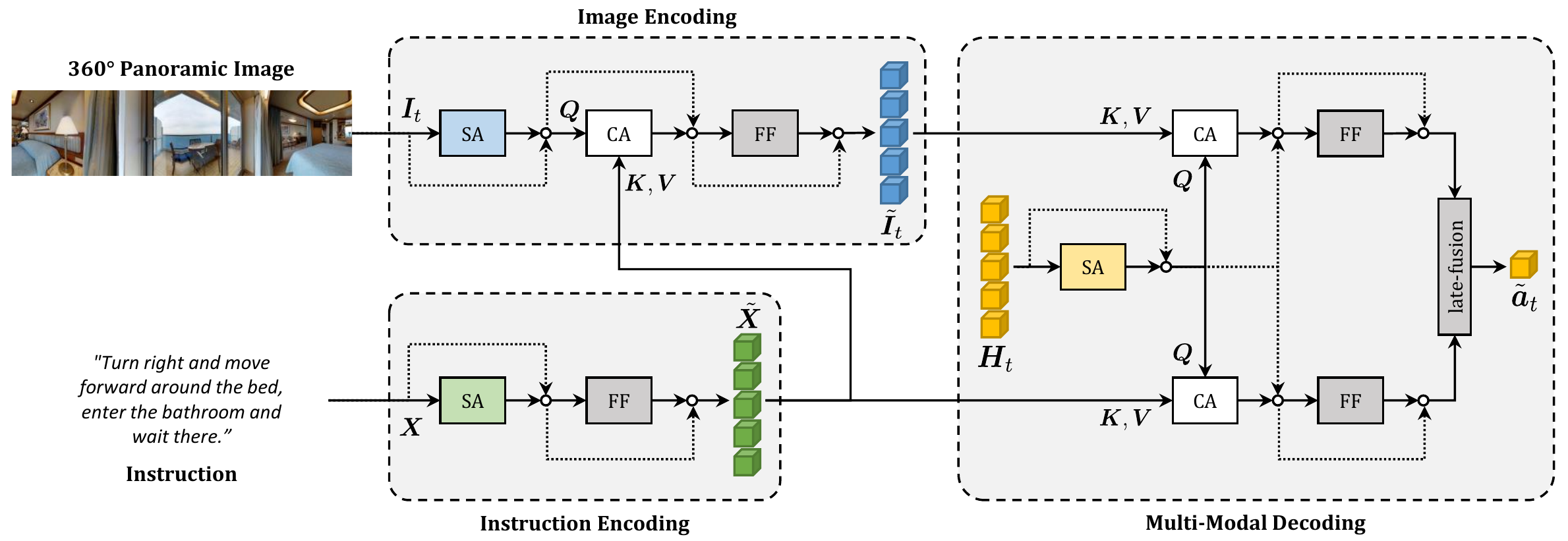}
    \caption{Overview of our approach. Our attention-based architecture for VLN builds upon three main blocks: an instruction encoder, an image encoder, and a multimodal decoder. SA, CA, and FF stand for self-attention, cross-attention, and feed-forward networks respectively. \rev{Dotted lines stand for residual connections between the results of the attention blocks and their inputs.} For sake of clarity, we omit layer normalization after each block}
    \label{fig:overview}
\end{figure*}

Our goal is to navigate unseen environments using low-level actions with the only help of natural language instructions and egocentric visual observations. To merge multimodal knowledge coming from the environment, we devise a \textbf{two-stage encoder}. \rev{In the first stage, we focus on encoding the instruction -- this step can be done once per episode as the natural language indication remains the same throughout the navigation. In the second stage, we use spatial attention to encode the visual observation and then employ the encoded instruction coming from the previous phase to enrich the agent representation of the surrounding environment.}
At each time step, the agent selects a move to progress towards the goal. To determine the next action, we fuse visuo-linguistic information with the history of actions via attention and build a \textbf{multimodal decoder} which merges the three modalities: actions, images, and text.
We then decode a probability distribution over a low-level output space in which possible actions are atomic moves like \textit{turn} or \textit{step ahead}. After a first phase in which we train the agent with classical imitation learning, we implement an \textbf{extrinsic reward} function to promote coherence between ground-truth and predicted trajectories.
%
We are the first, to the best of our knowledge, to build a VLN architecture without recurrence. Each component of our model is end-to-end trainable.
Our architecture is depicted in Fig.~\ref{fig:overview} and detailed next.

\subsection{Two-stage Encoder}
At the beginning of each navigation episode, the agent receives a natural language instruction $\{w_0, w_1, \dots, w_{n-1}\}$ of variable length $n$. The agent also perceives a panoramic $360\degree$ image of the surroundings $\bm{I}_t$ at each timestep $t$.
Our encoder consists of a single branch for each modality: text and images, and then employs attention to create a fused representation which specifically models the relevance of the source instruction into the visual observation.

\tit{Instruction Encoding}
To encode the textual instruction, we employ an attention mechanism with multiple heads, followed by a feed-forward network.
As a first step, we filter stop words and apply GloVe embeddings~\citep{pennington2014glove} to obtain a meaningful representation for each word.
We then apply the following transformation:
\begin{equation}
    \tilde{\bm{X}} = \text{LayerNorm}\left(\text{max}(0, \bm{X}\bm{W}_x + \bm{b}_x)\right),
    \label{eq:text_emb}
\end{equation}
where $\bm{X}$ is the GloVe embedding for the natural language instruction, $\bm{W}_x \in \mathbb{R}^{d_\text{GloVe} \times d_\text{model}}$ and $\bm{b}_x \in \mathbb{R}^{d_\text{model}}$ are learnable parameters, and $\text{LayerNorm}(\cdot)$ stands for layer normalization.
\rev{Since the instruction encoder has no recurrence, we must inject information about the relative position of the words in the sentence. Such information is added in the form of positional encoding to the input embeddings. The positional encodings have the same dimension as the embeddings, so that the two can be summed. We employ sine and cosine functions of different frequencies, in line with~\citep{vaswani2017attention}:
\begin{align}
\begin{split}
    PE_{(pos,2j)} = sin(pos / 10000^{2j/d_\text{model}}) \\
    PE_{(pos,2j+1)} = cos(pos / 10000^{2j/d_\text{model}})
\end{split}
\label{eq:pe}
\end{align}
where $pos$ is the position in the sequence and $j$ is the channel index.}
%
At this point we use multi-head attention to create a representation that models temporal dependencies inside the instruction. Multi-head attention is defined as:
\begin{align}
\begin{split}
    \text{MH}\left(\bm{Q},\bm{K},\bm{V}\right) = \text{Concat}\left(\bm{h}_1, \bm{h}_2, \dots, \bm{h}_h\right) \bm{W}^O  \\
    \text{with } \bm{h}_i = \text{Attention}\left(\bm{Q} \bm{W}_i^Q, \bm{K} \bm{W}_i^K, \bm{V} \bm{W}_i^V\right),
\end{split}
\label{eq:mh_att}
\end{align}
where 
\rev{$\bm{W}^Q_i \in \mathbb{R}^{d_\text{model} \times d_k}$,
$\bm{W}^K_i \in \mathbb{R}^{d_\text{model} \times d_k}$,
$\bm{W}^V_i \in \mathbb{R}^{d_\text{model} \times d_v}$,
and $\bm{W}^O \in \mathbb{R}^{hd_v \times d_\text{model}}$ denote learnable weight matrices, and the index $i$ stands for the $i^{\text{th}}$ head in the multi-head attention module. As also stated in our implementation details, $d_k = d_v = d_\text{model} / h$.}
\rev{In each head, we employ the scaled dot-product attention defined by~\cite{vaswani2017attention}:}
\begin{equation}
    \text{Attention}\left(\bm{Q}, \bm{K}, \bm{V}\right) = \text{softmax}\left( \frac{\bm{Q}\bm{K}^\top}{\sqrt{d_k}} \right) \bm{V}.
    \label{eq:att}
\end{equation}
The attention mechanism described by Eq.~\ref{eq:att} computes a weighted sum of the values ($\bm{V}$) basing on the similarity between the keys and the queries ($\bm{K}$ and $\bm{Q}$). In the self-attention, the same source sequence ($\tilde{\bm{X}}$ in this case) is employed to model the $(\bm{Q}, \bm{K}, \bm{V})$ triplet of Eq.~\ref{eq:mh_att}.
Following the attention layer, we place a feed-forward multilayer perceptron:
\begin{equation}
    \text{FF}\left(\tilde{\bm{X}}\right) = \text{max}\left(0, \tilde{\bm{X}}\bm{W}_1 + \bm{b}_1\right)\bm{W}_2 + \bm{b}_2,
    \label{eq:ffd}
\end{equation}
\rev{where $\bm{W}_1 \in \mathbb{R}^{d_\text{model} \times d_\text{ff}}$,
$\bm{W}_2 \in \mathbb{R}^{d_\text{ff} \times d_\text{model}}$,
$\bm{b}_1 \in \mathbb{R}^{d_\text{ff}}$,
$\bm{b}_2 \in \mathbb{R}^{d_\text{model}}$.}
At the end of this step, we obtain the attended representation for the current instruction $\tilde{\bm{X}} = \{\tilde{\bm{x}}_0, \tilde{\bm{x}}_1, \dots, \tilde{\bm{x}}_{n-1}\}$, that we use both during image encoding and in our multimodal decoder. 

\tit{Image Encoding}
As a first step, we discretize the $360\degree$ panoramic image of the surroundings $\bm{I}_t$ in 36 squared locations and we extract the corresponding visual features with a ResNet-152~\citep{he2016deep} trained on ImageNet~\citep{deng2009imagenet}. Each viewpoint covers $30\degree$ in the equirectangular image representing the agent surroundings, hence the image representation takes the form of a $3\times12$ grid. We then project visual features with a transformation similar to Eq.~\ref{eq:text_emb}, but instead of using sinusoidal positional encodings, we append a coordinate vector given by:
\begin{equation}
    \text{coord}_t = \left(\sin{\phi_t}, \cos{\phi_t}, \sin{\theta_t}\right),
    \label{eq:coord}
\end{equation}
where $\phi_t \in (-\pi, \pi]$ and $\theta_t \in [-\frac{\pi}{2}, \frac{\pi}{2}]$ are the heading and elevation angles for each viewpoint in the $3\times12$ grid relative to the agent position at timestep $t$.
We then apply multi-head self-attention according to Eq.~\ref{eq:mh_att} to help modeling concepts such as relative positions between objects. \rev{In this layer, the input sequence modeling $(\bm{Q}, \bm{K}, \bm{V})$ is composed by the features extracted from the 36 squared regions of $\bm{I}_t$.
}

After this step, we aim to create an image representation enriched with the textual concepts expressed by the attended instruction $\tilde{\bm{X}}$. We use cross-attention to achieve this goal, and employ $\tilde{\bm{X}}$ as keys and values for multi-head attention (Eq.~\ref{eq:mh_att}), \rev{while the queries come from the output of the previous self-attention layer. Using cross-attention, we enrich visual information with a weighted sum of the instruction tokens. From the resulting representation it is possible to draw concepts such as the \textit{tableness} or the \textit{redness} of an image region, given an instruction that refers to concepts such as \textit{table} or \textit{red}.} Finally, a feed-forward network as in Eq.~\ref{eq:ffd} is applied to obtain the attended visual observation $\tilde{\bm{I}}_t$. 

\subsection{Multimodal Decoder}
Our decoder predicts the next action to perform
among the following instructions: \textit{turn right/left $30\degree$}, \textit{tilt up/down}, \textit{step forward}, and \textit{end episode} -- to signal that it has reached the goal.

\tit{Contextual History for Action Decoding}
The first part of our decoder takes into account the history of past actions. While previous methods employ a recurrent neural network to keep track of previous steps \rev{(see for instance~\cite{anderson2018vision, ma2019self, wang2018reinforced})}, we explicitly model $\bm{H}_t = \{a_0, a_1, \dots, a_{t-1}\}$ as the set of actions performed before the current timestep $t$. Note that $a_0$ coincides with the \textit{\textless start\textgreater} token. We add sinusoidal positional encoding (Eq.~\ref{eq:pe}) to provide temporal information and apply multi-head self-attention to obtain an attended history representation $\tilde{\bm{H}}_t = \{\tilde{\bm{a}}_0, \tilde{\bm{a}}_1, \dots, \tilde{\bm{a}}_{t-1}\}$.

\tit{Late Fusion of Perception and Action}
At this point, $\tilde{\bm{H}}_t$ contains the relevant information regarding the action history of the navigation episode. However, this information must be enriched with the perception coming from the environment. We merge textual and visual information with $\tilde{\bm{H}}_t$ via attention, allowing mutual influence between perception and motion. We build two branches of multi-head cross-attention accepting respectively $\tilde{\bm{X}}$ and $\tilde{\bm{I}}_t$ as key/value pairs and using $\tilde{\bm{H}}_t$ as query. 
\rev{The image-action cross-attention is motivated by the fact that the agent needs to \textit{look around} before decoding the next action. Since $\tilde{\bm{I}}_t$ already contains information coming from the instruction, this cross-attention layer is sufficient to achieve decent results on the VLN task (as demonstrated by our ablation studies). However, we find out that adding a separate text-action cross-attention layer helps generalization in unseen environments.}
After this step, we concatenate the two representations and apply a FC layer to obtain the output sequence whose last element corresponds to $\tilde{\bm{a}}_t$. \rev{With this last layer, we perform a late fusion of visuo-linguistic information with the agent internal state (given by its previous history). It is worth noting that PTA also comprises an early fusion mechanism: the cross-attention between $\tilde{\bm{X}_t}$ and the attended visual input introduced in the Image Encoder. In our ablation study, we discuss the positive effects given by the early fusion and the late fusion mechanisms.}

\tit{Action Selection}
To select the next low-level action, we project the final representation $\tilde{\bm{a}}_t$ in a six-dimensional space corresponding with the agent locomotor space containing the following actions: \textit{turn right/left $30\degree$}, \textit{tilt up/down}, \textit{step forward}, and \textit{end episode}. The output probability distribution over the action space can therefore be written as:
\begin{equation}
    \bm{p}_t = \text{softmax}\left(\tilde{\bm{a}}_t \bm{W}_p + \bm{b}_p\right),
    \label{eq:softmax}
\end{equation}
where $\bm{W}_p \in \mathbb{R}^{d_\text{model} \times n_\text{actions}}$ and
$\bm{b}_p \in \mathbb{R}^{n_\text{actions}}$ are learned parameters ($n_\text{actions} = 6$).
During training, we sample the next action to perform $a_t$ from $\bm{p}_t$, while we select $a_t = \text{argmax}(\bm{p}_t)$ during evaluation and test.

\subsection{Training}
\label{subsec:training}
Our training setup includes two distinct objective functions. The first estimates the policy by imitation learning, while the second enforces similarity between the ground-truth and predicted trajectories via reinforcement learning.

\tit{Imitation Learning}
To approximate a good policy, we first train our agent using strong supervision. At each timestep $t$, the simulator outputs the ground-truth action $y_t$. In the low-level setup, the ground-truth action is the one that allows getting to the next target viewpoint in the minimum amount of steps.
%
In this phase, we aim to minimize the cross-entropy loss of the predicted distribution $\bm{p}_t$ \textit{w.r.t.} the ground-truth action $y_t$.

\tit{Extrinsic Reward}
After a first training phase with supervised learning, we finetune our agent using an extrinsic reward function.
Recently, \cite{magalhaes2019effective} propose to employ Dynamic Time Warping (DTW)~\citep{berndt1994using} to evaluate the trajectories performed by navigation agents. In particular, they define the \textit{normalized Dynamic Time Warping} (nDTW) as:
\begin{equation}
    \text{nDTW}(R, Q) = \exp\left(-\frac{\text{DTW}(R, Q)}{|R|\cdot d_{th}}\right),
\end{equation}
where $R$ and $Q$ are respectively the reference and the query paths, $|R|$ is the length of the reference path, and $d_{th}$ is the success threshold distance.
At each navigation step $t$, the agent receives a reward equal to the gain in terms of nDTW:
\begin{equation}
    R_t = \text{nDTW}(q_{0,\dots,t}, R) - \text{nDTW}(q_{0,\dots,t-1}, R).
\end{equation}
Additionally, we give an episode-level reward to the agent if it terminates the navigation within a success threshold distance $d_{th}$ from the goal, given by $R_s = \text{max}(0, 1 - d_{goal}/d_{th})$,
where $d_{goal}$ is the final distance between the agent and the target.
We can write our final reinforcement learning objective function as:
\begin{equation}
    L_{rl} = - \mathbb{E}_{a_t \sim \pi_\theta}\left[A_t\right].
\end{equation}
where the advantage function $A_t = R_t + R_s$.
Based on REINFORCE algorithm~\citep{williams1992simple}, we derive the gradient of our reward-based objective as:
\begin{equation}
    \nabla_\theta L_{rl} = -A_t \nabla \log \pi_\theta(a_t | s_t).
\end{equation}

\section{Low-level and High-Level Navigation}
\label{sec:ll_hl}
Section~\ref{sec:method} describes our approach to \textit{low-level} VLN.
Here, we discuss the main technical differences with the high-level counterpart and explain how PTA can switch from one setting to the other.
Differently from the low-level architectures, a high-level method aims to predict the next node to traverse in the navigation graph, as physical navigation takes place with a teleport mechanism. 
The choice at time step $t$ is done with a similarity measure between the agent internal state $\bm{s}_t$ and the appearance vector for the navigable locations $\bm{v}_t$.
This similarity function is normally mapped into a bilinear dot-product:
\begin{equation}
    \bm{p}_t = \text{softmax}\left(f(\bm{s}_t)^\top g(\bm{v}_t)\right)
\label{eq:bilinear}
\end{equation}
where $f(\cdot)$ and $g(\cdot)$ are generic transformations.


In principle, it is possible to substitute the final softmax classifier of a low-level architecture  (Eq.~\ref{eq:softmax}) with Eq.~\ref{eq:bilinear} and change the corresponding action space.
According to this observation, we can swap the action space of a model to test its adaptability to different navigation settings. 
While traditional approaches start from the hidden state of the recurrent policy to estimate the agent's internal state $\bm{s}_t$, we can derive it directly from $\tilde{\bm{a}}_t$:
\begin{equation}
    \bm{s}_t = \tilde{\bm{a}}_t \bm{W}_s + \bm{b}_s,
\end{equation}
where $\bm{W}_s$ and $\bm{b}_s$ are learned parameters.
As $\bm{v}_t$, we select the unattended visual features augmented with the coordinate vector described by Eq.~\ref{eq:coord}, and apply the following transformation:
\begin{equation}
    g(\bm{v}_t) = \text{max}\left(0, {\bm{v}}_t \bm{W}_v + \bm{b}_v\right),
\end{equation}
where $\bm{W}_v$ and $\bm{b}_v$ are learned parameters.

In our architecture, $\tilde{\bm{a}}_t$ can fit to represent any kind of information about the current navigation. This is because it can draw knowledge from the perceptual modalities and the history of past actions directly and without the bottleneck represented by a recurrent network. 
Our experiments on this subject (Sec.~\ref{subsec:r2r_res}) show that our model stands out from the literature in terms of adaptability. \rev{In other words, PTA can adapt to a different action space because it does not make any assumptions on the underlying simulating platform. Instead, our architecture relies on efficient visuo-linguistic fusion mechanisms designed to be agnostic towards the final action space. We will see that methods making stronger assumptions on the action space experience a larger drop in performance than PTA.}

\section{Experiments and Discussion}
\label{sec:experiments}
\begin{table*}[!t]
\small
\centering
\setlength{\tabcolsep}{.35em}
\resizebox{\linewidth}{!}{
\begin{tabular}{cclcccccccccccccccc}
\toprule
   & & & & \multicolumn{7}{c}{\textbf{Validation-Seen}} & & \multicolumn{7}{c}{\textbf{Validation-Unseen}} \\
  \cmidrule{5-11} \cmidrule{13-19}
  \textbf{\#}  & & \textbf{Method}
  & & NE $\downarrow$ & SR $\uparrow$ & OSR $\uparrow$ & SPL $\uparrow$ & CLS $\uparrow$ & nDTW $\uparrow$ & SDTW $\uparrow$ 
  & & NE $\downarrow$ & SR $\uparrow$ & OSR $\uparrow$ & SPL $\uparrow$ & CLS $\uparrow$ & nDTW $\uparrow$ & SDTW $\uparrow$ \\
\midrule
1 & & \cite{anderson2018vision} & &       
6.01 & 0.39 & 0.53 & - & - & - & - & &
7.81 & 0.22 & 0.28 & - & - & - & - \\
\midrule
2 & & PTA (pure IL, no extrinsic reward) & &
4.14 & 0.58 & 0.70 & 0.50 & 0.63 & 0.48 & 0.39 & &
6.44 & 0.39 & \textbf{0.49} & 0.32 & 0.48 & 0.32 & 0.24 \\
3 & & $-$ multi-modal decoder (only visual) & &
3.90 & 0.61 & 0.72 & 0.54 & 0.65 & 0.52 & 0.44 & &
6.56 & 0.36 & 0.46 & 0.29 & 0.47 & 0.32 & 0.22 \\
4 & & $-$ multi-modal decoder (only textual) & &
9.64 & 0.03 & 0.04 & 0.03 & 0.28 & 0.19 & 0.02 & & 
9.13 & 0.04 & 0.04 & 0.04 & 0.28 & 0.21 & 0.02 \\
5 & & $-$ early fusion (cross attention) & &
6.41 & 0.34 & 0.44 & 0.30 & 0.54 & 0.28 & 0.18 & & 
7.70 & 0.23 & 0.29 & 0.20 & 0.43 & 0.20 & 0.12 \\
6 & & $-$ action history (only last action) & &
5.40 & 0.42 & 0.54 & 0.36 & 0.55 & 0.39 & 0.27 & & 
7.19 & 0.22 & 0.31 & 0.18 & 0.41 & 0.26 & 0.12 \\
\midrule
7 & & $+$ data augmentation & &
\textbf{3.47} & \textbf{0.66} & \textbf{0.76} & 0.58 & 0.67 & 0.54 & 0.47 & &
\textbf{5.91} & 0.40 & 0.48 & 0.34 & 0.50 & 0.36 & 0.25 \\
8 & & $+$ extrinsic reward & &
3.58 & 0.65 & 0.74 & \textbf{0.59} & \textbf{0.69} & \textbf{0.60} & \textbf{0.50} & &
6.00 & \textbf{0.40} & 0.47 & \textbf{0.36} & \textbf{0.52} & \textbf{0.41} & \textbf{0.28} \\
\bottomrule 
\end{tabular}
}
\caption{Ablation study proving the effectiveness of our main modules. We also show that our model can be initialized using synthethic data augmentation and then finetuned with a limited set of refined data. Adding an extrinsic reward function further improves the performance in the final model.}
\label{table:ablation}
\end{table*}


\subsection{Experimental Setup}
\label{subsec:setup}
\tinytit{Datasets} 
In our experiments, we primarily test our architecture on the R2R dataset for VLN~\citep{anderson2018vision}. This dataset builds on the Matterport3D dataset of spaces~\citep{Matterport3D}, which contains complete scans of $90$ different buildings. The visual data is enriched with more than $7\,000$ navigation paths and $21\,000$ natural language instructions. The episodes are divided into a training set, two validation splits (\textit{validation-seen}, with environments that the agent has already seen during training, and \textit{validation-unseen}, containing only unexplored buildings), and a test set. The testing phase takes place in previously unseen environments and is accessible via a test-server with a public leaderboard.
While the instructions in R2R are quite long and complex (about $29$  words on average), navigation episodes usually involve a limited number of steps -- max $6$ steps for high-level action space and max $23$ steps for the low-level setup. In the R4R dataset, ~\cite{sotp2019acl}, merge the paths in R2R to create a more complex and challenging setup. Episodes become considerably longer, pushing the traditional approaches to their limits and testing their generalizability to arbitrary long instructions and more complex trajectories.

\tit{Evaluation Metrics}
In line with previous literature, we mainly focus on four metrics. NE (Navigation Error) measures the mean distance from the goal and the stop point. SR (Success Rate) is the fraction of episodes concluded within a threshold distance from the target -- 3 meters for all of the previous papers on the subject.
OSR (Oracle SR) represents the SR that the agent would achieve if it received an oracle stop signal when passing within the threshold distance from the goal, while SPL (SR weighted by inverse Path Length) penalizes navigation episodes that deviate from the shortest path to the goal. SPL is accredited to be the most reliable metric on the R2R dataset~\citep{anderson2018evaluation}, as it strongly penalizes exhaustive exploration and search methods like beam search. Recently, \cite{sotp2019acl} propose to use Coverage weighted by Length Score (CLS) to replace SR for generic navigation trajectories, as this metric is also sensitive to intermediate nodes in the reference path. Additionally, \cite{magalhaes2019effective} propose Dynamic Time Warping (DTW) and derived metrics (Normalized DTW and Success weighted by normalized DTW) to measure the similarity between reference and predicted paths. These three last metrics are more meaningful on the R4R dataset than SR and SPL~\citep{sotp2019acl}.

\tit{Implementation Details}
\rev{In the instruction encoder, $d_\text{GloVe} = 300$. In each component of our model, we project the input features into a $d_\text{model}$-dimensional space, with $d_\text{model} = 512$. For multi-head attention, we employ $h = 8$ heads, thus $d_k = d_v = d_\text{model} / h = 64$. The internal representation of feed-forward networks has size $d_\text{ff} = 2048$.} After each sub-module, we add a residual connection followed by layer normalization. We also apply dropout~\citep{srivastava2014dropout} with drop probability $p=0.1$ after each linear layer. During training, we use Adam optimizer~\citep{kingma2015adam} with learning rate $10^{-4}$, we set the batch size to $32$ and reduce the learning rate by a factor $10$ if the SPL on the validation unseen split does not improve for $5$ consecutive epochs. We stop the training after $30$ epochs without improvement on the same metric. When finetuning using REINFORCE, we set the initial learning rate to $10^{-7}$. 

\subsection{Ablation Study}
\label{subsec:ablation}
In our ablation study, we experimentally validate the importance of each module in our architecture. First, we ablate multi-modality in our decoder and we do not apply late fusion before decoding the next action. In a second experiment, we remove cross-attention between visual and lingual information in the encoder.
Finally, we show the impact of synthetic data augmentation~\citep{fried2018speaker} and the role of REINFORCE. Results are shown in Table~\ref{table:ablation} and discussed below.

\tit{Multimodal Decoder}
In our first ablation study, we use only one of the two decoder branches at the time, and we do not perform late fusion between lingual and visually-grounded information.
When removing the textual branch (Table~\ref{table:ablation}, line 3), our agent performs worse on unseen environments, hence losing potential in terms of generalization. When removing the visual modality, our PTA agent is blinded and can only count on the natural language instruction. This setup leads to success only when the instruction does not involve references to objects or visual properties of the environment -- a nearly empty subset of the dataset. Indeed, the metrics for our \textit{blind} agent are extremely low, and they do not vary between seen and unseen environments  (Table~\ref{table:ablation}, line 4). This result is meaningful in light of recent studies proving that some single-modality agents perform better than their multimodal version by removing the visual perception and overfitting on dataset biases~\citep{Thomason2018ShiftingTB}.

\tit{Early Fusion of Textual and Visual Perception}
As a second experiment, we remove the early fusion mechanism, namely the cross-attention layer between the textual and visual branches of our encoder, to check its contribution. If this fusion layer is redundant, we expect that the late fusion stage will compensate for the loss. Instead, we experience a drop in performance: $-12\%$ in SPL in unseen environments (Table~\ref{table:ablation}, line 5).
We thus prove the importance of early textual and visual fusion in our architecture for VLN.

\tit{Contextual History for Action Decoding}
$\bm{H}_t$ stores past actions as a series of one hot vectors, and it is extremely helpful to model navigation history. It acts as a sort of memory for the agent, so that it knows what actions have already been made. A similar trick in LSTM-based VLN consists in adding the last action as input to the policy RNN at each step. In our model, removing $\bm{H}_t$ and using only the last action (losing all the history) causes a drop in performance: $-14\%$ and $-17\%$ on SPL and SR respectively for the Val-Unseen split (Table~\ref{table:ablation}, line 6).

\tit{Data Augmentation}
In line with previous literature, we find the use of additional synthetic instructions useful to initialize our agent. The synthetic training set was provided by \cite{fried2018speaker} using a \textit{Speaker} module. After a first training with the full set of instructions (synthetic \textit{and} human-generated), we finetune using \textit{only} the original R2R train set. Results are reported in Table~\ref{table:ablation}, line 7.

\tit{Extrinsic Reward}
While imitation learning allows approximating a good policy, there is still room for improvement via reinforcement learning. \cite{wang2018reinforced} were the first to use REINFORCE in the context of VLN to refine their navigation policy based on cross-modal matching.
In line with them, we find REINFORCE beneficial for our model: our final agent sticks more closely to the reference trajectory and penalizes overlong navigations (Table~\ref{table:ablation}, line 8).

\subsection{Results on R2R}
\label{subsec:r2r_res}
In our experiments on the R2R dataset~\citep{anderson2018vision}, we test the ability of our agent to navigate unseen environments in light of previously unseen natural language instructions. The main test-bed for this experiment is represented by the R2R evaluation leaderboard, which is publicly available online.

\tit{Comparison with SOTA}
In Table~\ref{table:test}, we report our results on the R2R test set, together with the results achieved by other state-of-the-art architectures on VLN. Other methods that operate in the \textit{low-level action space} are the sequence-to-sequence architecture proposed by \cite{anderson2018vision}, the RPA model using a mixture of model-free and model-based reinforcement learning~\citep{wang2018look}, and the recurrent architecture with dynamic convolutional filters proposed by \cite{landi2019embodied}. Our method overcomes the state-of-the-art on low-level VLN by a large margin ($5\%$ in terms of SPL and SR).

Although a direct comparison between the two settings is not feasible, we notice that PTA performs better than some high-level architectures in terms of SPL. Notably, we achieve this result without making any assumption on the underlying simulating platform and decoding a longer sequence of atomic moves, instead of target viewpoints. Moreover, high-level architectures can often count on efficient graph-search methods (impractical when dealing with continuous controls) to decode the final trajectory, and on additional modules that are not present in our method. While these are effective for high-level VLN, their generalizability to a low-level setup, closer to real-world application, is yet to be tested.

\begin{table}[t!]
\centering
\footnotesize
\setlength{\tabcolsep}{.4em}
\begin{tabular}{lccccc}
\toprule
& & \multicolumn{4}{c}{\textbf{Test (Unseen)}} \\
\cmidrule{3-6}
\textbf{Low-level Methods} & &
NE $\downarrow$ & SR $\uparrow$ & OSR $\uparrow$ & SPL $\uparrow$ \\
\midrule
Random &           
& 9.77 & 0.13 & 0.18 & 0.12   \\
\cite{anderson2018vision} &
& 7.85 & 0.20 & 0.27 & 0.18   \\
\cite{wang2018look} &
& 7.53 & 0.25 & 0.33 & 0.23   \\
\cite{landi2019embodied} &
& 6.55 & 0.35 & 0.45 & 0.31 \\
\midrule
\textbf{PTA} &
& \textbf{6.17} & \textbf{0.40} & \textbf{0.47} & \textbf{0.36} \\
\bottomrule
\end{tabular}
\setlength{\tabcolsep}{.49em}
\begin{tabular}{lccccc}
\toprule
& & \multicolumn{4}{c}{\textbf{Test (Unseen)}} \\
\cmidrule{3-6}
\textbf{High-level Methods} & &
NE $\downarrow$ & SR $\uparrow$ & OSR $\uparrow$ & SPL $\uparrow$ \\
\midrule
\cite{fried2018speaker} &
& 6.62 & 0.35 & 0.44 & 0.28 \\
\cite{ma2019self} &
& 5.67 & 0.48 & 0.59 & 0.35 \\
\cite{wang2018reinforced} &
& 6.01 & 0.43 & 0.51 & 0.35 \\
\cite{ma2019regretful} &
& 5.69 & 0.48 & 0.56 & 0.40 \\
\cite{ke2017tactile} &
& 5.14 & 0.54 & \textbf{0.64} & 0.41 \\
\cite{tan2019learning} &
& 5.23 & 0.51 & 0.59 & 0.47 \\
\cite{li2019efficient} &
& \textbf{4.53} & \textbf{0.57} & 0.63 & \textbf{0.53} \\
\bottomrule
\end{tabular}
\caption{Results on the R2R test server for low-level (top) and high-level (bottom) methods. We chose the best version of each model basing on SPL.}
\label{table:test}
\end{table}

\begin{figure*}[t!]
\centering
    \includegraphics[width=.6\linewidth]{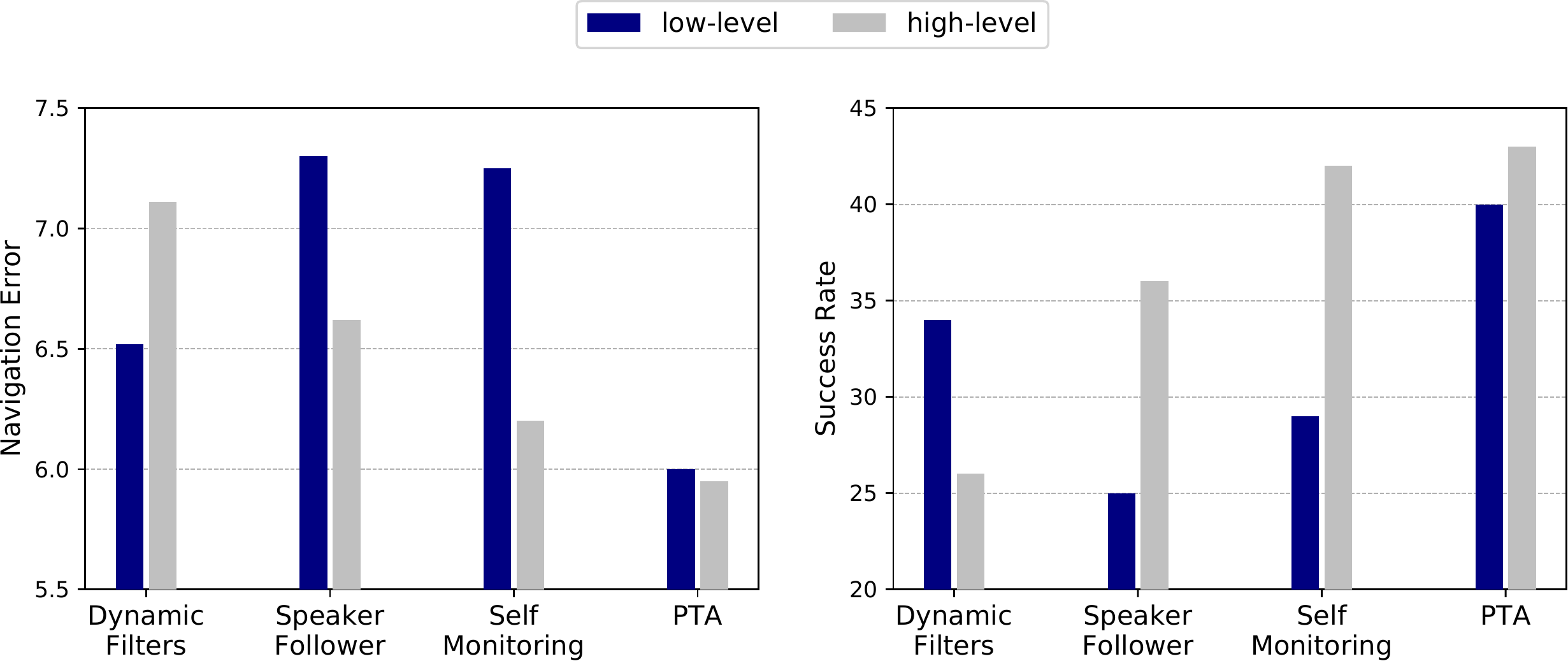}
    \caption{Visualization of the navigation error (left) and success rate (right) on the R2R val-unseen split. A larger difference between the blue and gray bars denotes a lower degree of adaptability. The metric gap is reduced when using PTA}
    \label{fig:plot}
    \vspace{+.2cm}
\end{figure*}

\tit{Switching from Low-level to High-level}
\rev{Our second experiment on R2R aims to test the effects of retraining existing models after switching their final action spaces (from high-level to low-level and vice-versa).
To that end, we change the final classifier of PTA as described in Section~\ref{sec:ll_hl}. In this new setting, the output of the action decoder becomes a probability distribution over the adjacent nodes of the navigation graph. Once the agent decides where to go, the displacements are made automatically and there is no need to decode lower-level actions such as rotations. We train PTA from scratch in this setup, without any further hyperparameter tuning.}
In Table~\ref{table:pta_ll_hl} we detail the full set of metrics obtained using PTA with the high-level classifier, and compare with the model incorporating the low-level control system.
The small gap between the metrics in the two setups suggests that PTA \rev{does not take any particular advantage from the underlying action space}. Of course, metrics that directly evaluate the final trajectory (like DTW-based metrics) benefits from using high-level actions with automatic oracle displacements.

In principle, every model should exhibit a decent level of elasticity towards different locomotor settings. In practice, we find out that architectural choices that strongly help high-level VLN often end up hindering the other setup. \rev{This is especially true when the agent exploits high-level reasoning and makes strong assumptions on the nature of the underlying simulator.} As a result, current high-level methods experience a drop in performance when adopting a simple, atomic action space (see Figure~\ref{fig:plot}). \rev{PTA, instead, does not rely on such assumptions and builds on more efficient modules to merge multi-modal information entailed in the VLN task.}
The plots in Figure~\ref{fig:plot} show that our model exhibits far greater flexibility to the final action space than other architectures.
The considerably narrow step between the blue and the gray bars (representing the low-level and the high-level actions spaces respectively) denotes that a change in the final action space does not prevent PTA from reaching its goal.
We compare with the Speaker-Follower~\citep{fried2018speaker} and the Self-Monitoring agent~\citep{ma2019self} from the high-level setup, \rev{which experience a sizeable loss in performance}. In fact, results drop of $11\%$ and $13\%$ respectively in terms of SR when adapted for low-level use. We also compare PTA with a recurrent architecture exploiting dynamic convolution~\citep{landi2019embodied} from the low-level category. \rev{The lower degree of adaptability shown by this competitor is motivated by the fact that it operates a strong compression on the visual input basing on the current instruction. In this step, much information that could ease high-level action selection is lost.}

\begin{table}[t!]
\footnotesize
\centering
\setlength{\tabcolsep}{.35em}
\begin{tabular}{lcccccccc}
\toprule
\textbf{Method}  &
& NE $\downarrow$ & SR $\uparrow$ & OSR $\uparrow$ & SPL $\uparrow$ & CLS $\uparrow$ & nDTW $\uparrow$ & SDTW $\uparrow$\\
\midrule
\textbf{PTA} \textit{low-level} &
& 6.00 & 0.40 & 0.47 & 0.36 & 0.52 & 0.41 & 0.28 \\ 
\textbf{PTA} \textit{high-level} &
& 5.95 & 0.43 & 0.49 & 0.39 & 0.53 & 0.53 & 0.35  \\
\bottomrule
\end{tabular}
\caption{Comparison between the low-level and the high-level version of PTA. On all the metrics, a small gap denotes high adaptaility. DTW-based metrics highly benefits from the use of a high-level action space}
\label{table:pta_ll_hl}
\end{table}

\begin{figure*}[t]
    \centering
    \includegraphics[width=\linewidth]{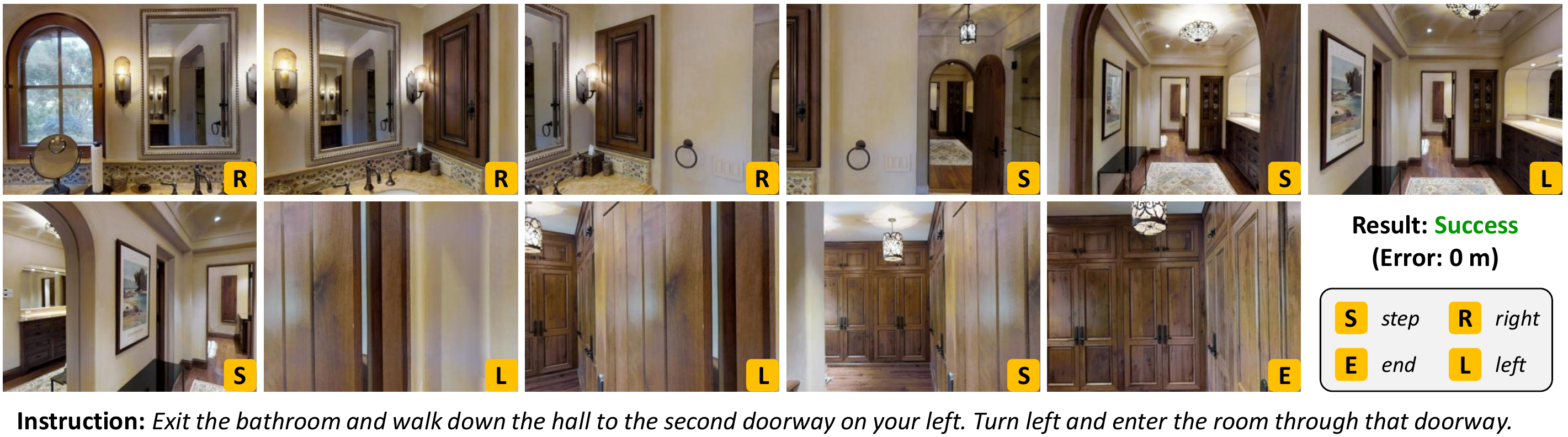}
    \caption{Navigation episode from the R2R unseen validation split. For each step, we report the agent first-person point of view and the next predicted action (from left to right, top to bottom)}
    \label{fig:results}
\end{figure*}

\begin{table*}[t]
\small
\centering
\setlength{\tabcolsep}{.35em}
\resizebox{\linewidth}{!}{
\begin{tabular}{lcccccccccccccccc}
\toprule
   & & \multicolumn{7}{c}{\textbf{R4R Validation-Seen}} & & \multicolumn{7}{c}{\textbf{R4R Validation-Unseen}} \\
  \cmidrule{3-9} \cmidrule{11-17}
  \textbf{Method}
  & & PL $\downarrow$ & NE $\downarrow$ & SR $\uparrow$ & SPL $\uparrow$ & CLS $\uparrow$ & nDTW $\uparrow$ & SDTW $\uparrow$ 
  & & PL $\downarrow$ & NE $\downarrow$ & SR $\uparrow$ & SPL $\uparrow$ & CLS $\uparrow$ & nDTW $\uparrow$ & SDTW $\uparrow$ \\
\midrule
\cite{landi2019embodied} & &       
11.9 & 5.74 & 0.51 & 0.39 & 0.50 & 0.38 & 0.24 & &
\textbf{9.98} & 9.03 & 0.20 & 0.11 & 0.33 & 0.19 & 0.06 \\
\textbf{PTA \textit{low-level}} & &       
\textbf{11.9} & \textbf{5.11} & \textbf{0.57} & \textbf{0.45} & \textbf{0.52} & \textbf{0.42} & \textbf{0.29} & &
10.2 & \textbf{8.19} & \textbf{0.27} & \textbf{0.15} & \textbf{0.35} & \textbf{0.20} & \textbf{0.08} \\
\midrule
\cite{fried2018speaker} & &
\textbf{15.4} & 5.35 & 0.52 & 0.37 & 0.46 & - & - & &
19.9 & 8.47 & 0.24 & \textbf{0.12} & 0.30 & - & - \\
RCM \textit{goal oriented}~\citep{sotp2019acl} & &
24.5 & 5.11 & 0.56 & 0.32 & 0.40 & - & - & &
32.5 & 8.45 & \textbf{0.29} & 0.10 & 0.20 & - & - \\
RCM \textit{fidelity oriented}~\citep{sotp2019acl} & &
18.8 & 5.37 & 0.53 & 0.31 & 0.55 & - & - & &
28.5 & \textbf{8.08} & 0.26 & 0.08 & 0.35 & - & - \\
\textbf{PTA \textit{high-level}}& &       
16.5 & \textbf{4.54} & \textbf{0.58} & \textbf{0.39} & \textbf{0.60} & \textbf{0.58} & \textbf{0.41} & &
\textbf{17.7} & 8.25 & 0.24 & 0.10 & \textbf{0.37} & \textbf{0.32} & \textbf{0.10} \\
\bottomrule 
\end{tabular}
}
\caption{Results on the R4R validation splits. Our model is the new state-of-the-art on the two splits in both of its versions -- \textit{low-level} and \textit{high-level}. Note that, since the trajectories can bind and return on the agent previous steps, CLS and nDTW are the more indicative metrics. Metrics with `-' were not reported in the original papers.}
\label{table:r4r}
\end{table*}

To conduct this experiment we adjust the codes from \cite{landi2019embodied} and \cite{ma2019self}, which are publicly available online, and report the results in the paper for \cite{fried2018speaker}. We choose the Speaker-Follower and the Self-Monitoring agents because they are flexible frameworks by design, and for this reason they are the most suitable models among their high-level peers for this comparison. \rev{We believe that the findings and insights provided in this experiment will motivate further experiments in this direction, and help to unravel the main reasons of improvements in new architectures for VLN.}


\tit{Qualitative Results}
In Fig.~\ref{fig:results}, we report a qualitative result from the R2R val-unseen set. Remarkably, PTA is able to ground concepts such as ``\textit{the second doorway on your left}'' and terminates the navigation episode successfully. Since our agent operates in a low-level setup, it needs to orientate towards the next viewpoint before stepping ahead, making the decoding phase more challenging.

\subsection{Results on R4R}
\label{subsec:r4r_res}
R4R~\citep{sotp2019acl} builds upon R2R and aims to provide an even more challenging setting for embodied navigation agents. 
While navigation in R2R is usually direct and takes the shortest path between the starting position and the goal viewpoint, trajectories in R4R may bend and return on the agent's previous steps. This change calls for adaptation in evaluation metrics: SPL and SR are now less indicative because the agent might stop close the goal in the first half of the navigation and still fail to complete the second part. In this sense, an important role is played by recently proposed metrics: CLS~\citep{sotp2019acl} and nDTW~\citep{magalhaes2019effective} take into account the agent's steps and are sensitive to intermediate errors in the navigation path.
For this reason, these last metrics are more meaningful when evaluating navigation agents on R4R.

\tit{Comparison with SOTA}
In this experiment, we compare PTA with other state-of-the-art architectures for VLN and report the results in Table~\ref{table:r4r}. In the low-level setup, we compare to the recurrent architecture with dynamic convolution proposed by \cite{landi2019embodied}. Results show that our approach performs better on all of the main metrics. In particular, a lower NE and a higher CLS indicate that our agent tends to get closer to the goal while sticking to the natural language instruction better than the competitor. 
We also report the results obtained by our model incorporating the high-level decision space. We compare with Speaker-Follower~\citep{fried2018speaker} and RCM~\citep{wang2018reinforced}, as implemented in \citep{sotp2019acl}. PTA performs better than its high-level competitors on the majority of the metrics. In particular, the higher CLS score shows that PTA can generally select a path that follows the instruction better than the competitors. When considering the reference metrics proposed for R4R~\citep{sotp2019acl}, our architecture achieves the best results on both the setups. 

\section{Conclusion}
\label{sec:conclusion}
In this paper, we have presented \textit{Perceive, Transform, and Act} (PTA), the first fully-attentive model for VLN. In particular, we tackle the challenging task of low-level VLN, in which high-level information about the environment is no longer accessible to the agent.  
We show that previous work on high-level VLN suffers from low flexibility and experiences a drop in performance when adapted for low-level use, while our agent naturally adapts to the other action space.
\revmin{These results suggest that boosts in performance observed in high-level VLN may be due to the use of a simpler action space, and encourage further research in this direction.}
\revmin{Our architectural choices allow for a significant boost in performance: PTA achieves good results on low-level VLN, and when testing on the recently proposed R4R dataset, PTA achieves promising results in both the setups.}


\section*{Acknowledgement}
This work has been supported by ``Fondazione di Modena'' and by the national project ``IDEHA: Innovation for Data Elaboration in Heritage Areas'' (PON ARS01\_00421), cofunded by the Italian Ministry of University and Research.

\bibliographystyle{model2-names}
\bibliography{refs}

\end{document}